\documentclass[10pt,twocolumn,letterpaper]{article}

\usepackage{iccv}
\usepackage{times}
\usepackage{epsfig}
\usepackage{graphicx}
\usepackage{amsmath}
\usepackage{amssymb}
\usepackage{pifont}
\usepackage{multirow}
\usepackage{array}
\usepackage[dvipsnames]{xcolor}
\usepackage[title]{appendix}
\usepackage{subcaption}
\usepackage{caption}
\usepackage[misc]{ifsym}
\newcolumntype{L}[1]{>{\raggedright\let\newline\\\arraybackslash\hspace{0pt}}m{#1}}
\newcolumntype{C}[1]{>{\centering\let\newline\\\arraybackslash\hspace{0pt}}m{#1}}
\newcolumntype{R}[1]{>{\raggedleft\let\newline\\\arraybackslash\hspace{0pt}}m{#1}}

\usepackage[pagebackref=true,breaklinks=true,letterpaper=true,colorlinks,bookmarks=false]{hyperref}

\iccvfinalcopy % *** Uncomment this line for the final submission

\def\iccvPaperID{2796} % *** Enter the ICCV Paper ID here

\ificcvfinal\fi
\begin{document}

\providecommand{\huiyu}[1]{{\protect\color{ForestGreen}{\bf [Huiyu: #1]}}}

%%%%%%%%% TITLE
\title{Semantic-Aware Knowledge Preservation for\\ Zero-Shot Sketch-Based Image Retrieval}

\author{Qing Liu\textsuperscript{1}, Lingxi Xie\textsuperscript{1,2,\Letter}, Huiyu Wang\textsuperscript{1}, Alan L. Yuille\textsuperscript{1}\\
\textsuperscript{1}Johns Hopkins University\quad\textsuperscript{2}Noah's Ark Lab, Huawei Inc.\\
{\tt\small qingliu@jhu.edu, 198808xc@gmail.com, huiyu@jhu.edu, alan.l.yuille@gmail.com}\\
}

\maketitle
%\thispagestyle{empty}

%%%%%%%%% ABSTRACT
\begin{abstract}
Sketch-based image retrieval (SBIR) is widely recognized as an important vision problem which implies a wide range of real-world applications. Recently, research interests arise in solving this problem under the more realistic and challenging setting of zero-shot learning. In this paper, we investigate this problem from the viewpoint of domain adaptation which we show is critical in improving feature embedding in the zero-shot scenario. Based on a framework which starts with a pre-trained model on ImageNet and fine-tunes it on the training set of SBIR benchmark, we advocate the importance of preserving previously acquired knowledge, e.g., the rich discriminative features learned from ImageNet, to improve the model's transfer ability. For this purpose, we design an approach named Semantic-Aware Knowledge prEservation (SAKE), which fine-tunes the pre-trained model in an economical way and leverages semantic information, e.g., inter-class relationship, to achieve the goal of knowledge preservation. Zero-shot experiments on two extended SBIR datasets, TU-Berlin and Sketchy, verify the superior performance of our approach. Extensive diagnostic experiments validate that knowledge preserved benefits SBIR in zero-shot settings, as a large fraction of the performance gain is from the more properly structured feature embedding for photo images. Code is available at: \url{https://github.com/qliu24/SAKE}.
\end{abstract}
\vspace{-0.5cm}
%%%%%%%%% BODY TEXT
\section{Introduction}

\newcommand{\figurewidth}{8cm}
\begin{figure}[!t]
\centering
\includegraphics[width=\figurewidth]{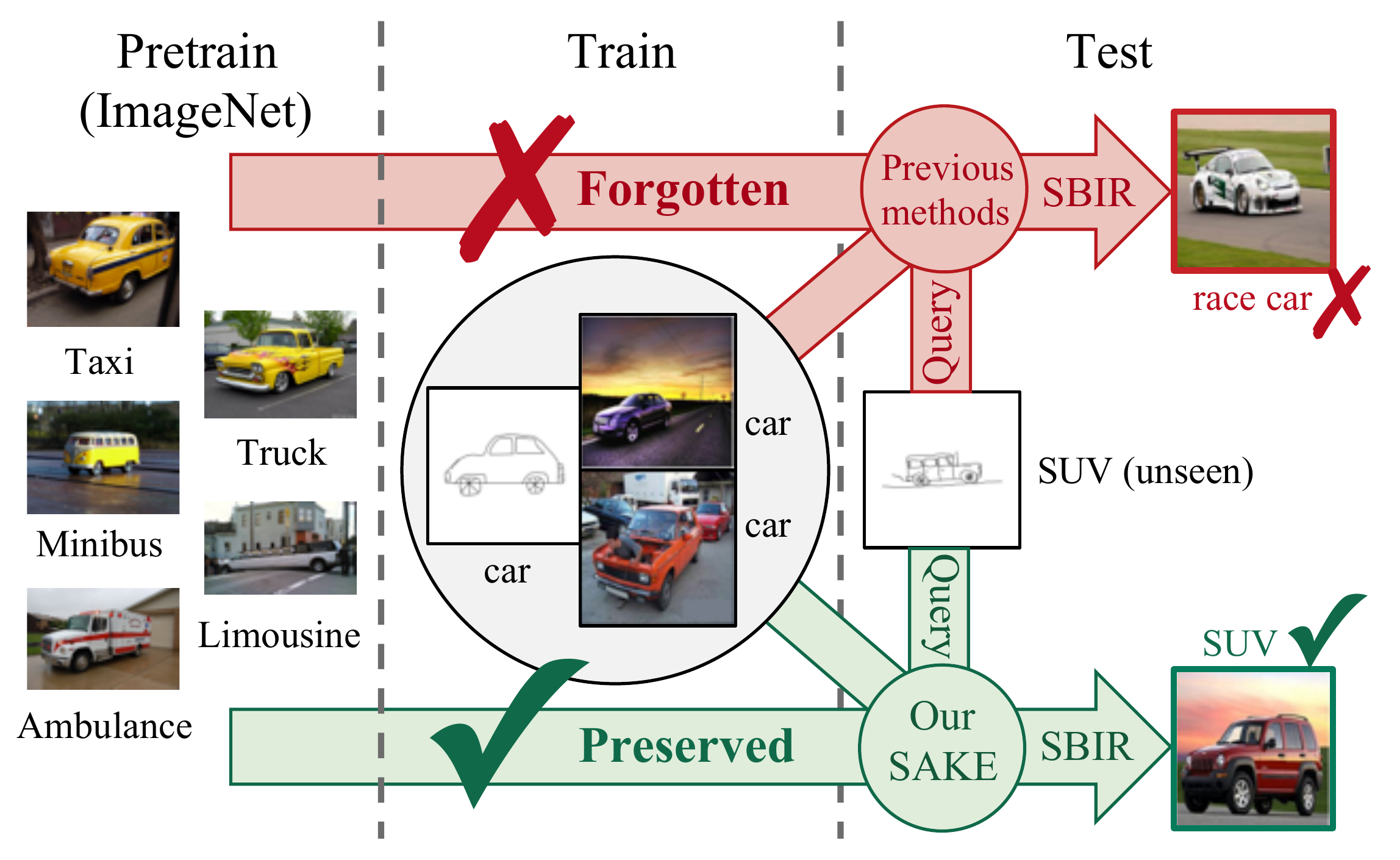}
\vspace{-0.2cm} \caption{An illustration of the ZS-SBIR task and our model. Catastrophic forgetting is harmful, especially in zero-shot settings. Our Semantic-Aware Knowledge prEservation (SAKE) preserves original domain knowledge of rich visual features (\emph{e.g.}, visual details of different subtypes of cars) which helps distinguishing the right photo candidates (\emph{e.g.}, SUV) from distractors (\emph{e.g.}, race car) in the \emph{unseen} classes.}
\vspace{-0.2cm} \label{Fig:Concept}
\end{figure}

Sketch-based image retrieval (SBIR) is an important, application-driven problem in computer vision \cite{eitz2011sketch, hu2011bag, eitz2010evaluation, kato1992sketch}. Given a hand-drawn sketch image as the query and a large database of photo images as the gallery, the goal is to find relevant images, {\em i.e.}, those with similar visual contents or the same object category, from the gallery. The most important issue of this task lies in finding a shared feature embedding for cross-modality data, which requires mapping each sketch and photo image to a high-dimensional vector in the feature space. In recent years, with the rapid development of deep learning, researchers have introduced deep neural networks into this field~\cite{lu2018learning, zhang2018generative, liu2017deep, yu2017sketch, song2017deep, sketchy2016, qi2016sketch, zhang2016sketchnet, yu2016sketch, wang2015sketch}. In the conventional setting, it is assumed that training and testing images are from the same set of object categories, in which scenario existing approaches achieved satisfying performance \cite{lu2018learning}. However, in real-world applications, there is no guarantee that the training set covers all object categories in the gallery at the application stage.
% Afterward, an efficient data structure, {\em e.g.}, the inverted index, is built to accomplish the task. One extreme case is that the test data is from a completely different set of categories that the model has never seen during training.

This paper investigates this more challenging setting in an extreme case. This setting is named zero-shot sketch-based image retrieval (ZS-SBIR), which assumes that classes in the target domain are unseen during the training stage. The goal of this setting is to test the model's ability to adapt learned knowledge to an unknown domain. Experiments show that existing SBIR models generally produce low accuracy in this challenging setting \cite{dutta2019semantically, shen2018zero, kiran2018zero}, possibly because they over-fitted the source domain and meanwhile being unaware of the \emph{unseen} categories. To tackle this problem, we call for a model to simultaneously solve the problems of object recognition, cross-modal matching, and domain adaptation.

An important observation of ours is that the unsatisfying performance in zero-shot learning is closely related to the catastrophic forgetting phenomenon~\cite{kirkpatrick2017overcoming,french1999catastrophic} during sequential learning, \emph{i.e.}, the task-specific fine-tuning process. All existing ZS-SBIR models fine-tune an ImageNet pre-trained model with mixed loss functions, \emph{e.g.}, a softmax-based term to distinguish different classes and a reconstruction loss term to learn shared image representations \cite{dutta2019semantically}. However, catastrophic forgetting implies that the previously acquired domain knowledge, {\em e.g.}, rich discriminative features learned from ImageNet, is mostly eliminated from the model during the fine-tuning process if it is not relevant to the new task. This results in the features being over-fitted to the limited data in the source domain and thus less capable of effectively representing and distinguishing samples in the target domain which contains unseen categories (an example is given in Figure~\ref{Fig:Concept}). To verify this, we fine-tune an ImageNet pre-trained AlexNet~\cite{krizhevsky2012imagenet} using data in the new source domain. We then fix the network and use the \texttt{fc7} features to train a linear classifier again on ImageNet, \emph{i.e.}, the original domain. Before fine-tuning, the model reports a classification accuracy of $56.29\%$, while this number drops to $45.54\%$ afterward. This experiment verifies that the model forgets part of the knowledge learned from ImageNet during the fine-tuning process.

Based on this observation, we propose a novel framework named Semantic-Aware Knowledge prEservation (SAKE), which aims at maximally preserving previously acquired knowledge during fine-tuning. SAKE does not require the access to the original ImageNet data but instead designs an auxiliary task to approximately map each image in the training (fine-tuning) set to the ImageNet semantic space. More specifically, the approximation is made during a teacher-student optimization process, in which the pre-trained model on ImageNet, with all parameters fixed, provides a teacher signal. We also use semantic information to refine the teacher signal to provide better supervision. An illustration of our motivation is shown in Figure~\ref{Fig:Concept}.

Following convention, we perform experiments on two popular SBIR datasets, namely, the TU-Berlin dataset~\cite{eitz2012hdhso} and the Sketchy dataset~\cite{sketchy2016}. Results verify the effectiveness of SAKE in boosting ZS-SBIR compared to state-of-the-art methods, and these gains also persist after we binarize the image features using iterative quantization (ITQ) \cite{gong2013iterative}. In addition, SAKE requires moderate extra computations and little memory during training and uses no extra resources in the testing stage. This eases its application in real-world scenarios.

The remainder of this paper is organized as follows. Section~\ref{sec:relatedwork} briefly introduces related works. Section~\ref{sec:approach} describes the problem setting and our solution. After experiments are shown in Section~\ref{sec:experiments}, we conclude this work in Section~\ref{sec:conclusions}.

%------------------------------------------------------------------------
\section{Related Work}
\label{sec:relatedwork}
\paragraph{SBIR and ZS-SBIR.}
The fundamental problem of SBIR task is to learn a shared representation to bridge the modality gap between the hand-drawn sketches and the real photo images.
Early works employed hand-crafted features to represent the sketches and matched them with the edge maps extracted from the photo images using different variants of the Bag-Of-Words model~\cite{saavedra2015sketch, hu2013performance, eitz2011sketch, hu2011bag, eitz2010evaluation}. In recent years, the deep neural networks (DNNs) were introduced into this field~\cite{lu2018learning, zhang2018generative, liu2017deep, yu2017sketch, song2017deep, sketchy2016, qi2016sketch, zhang2016sketchnet, yu2016sketch, wang2015sketch}. First proposed by~\cite{shen2018zero} and followed by~\cite{kiran2018zero, dutta2019semantically}, studies of SBIR in the zero-shot setting arose. To encourage the transfer of the learned cross-modal representations from the source domain to the target domain, ZS-SBIR works leveraged side information in semantic embeddings~\cite{dutta2019semantically, shen2018zero} and employed deep generative models, such as generative adversarial networks (GANs)~\cite{dutta2019semantically} and variational auto-encoders (VAEs)~\cite{shen2018zero, kiran2018zero}.
\vspace{-0.4cm}
\paragraph{Catastrophic Forgetting.} 
When a pre-trained model is fine-tuned to another domain or a different task, it tends to lose the ability to do the original task in the original domain. This phenomenon is called catastrophic forgetting~\cite{goodfellow2013empirical, french1999catastrophic, mccloskey1989catastrophic} and observed in training neural networks. Incremental learning methods~\cite{mao2015learning, cauwenberghs2001incremental, polikar2001learn++,thrun1996learning,schlimmer1986case} adapted models to gradually available data and required overcoming catastrophic forgetting. \cite{kirkpatrick2017overcoming} proposed to selectively slow down learning on the weights that are important for old tasks. Later,~\cite{li2018learning, shmelkov2017incremental} proposed to mimic the original model's response for old tasks at the fine-tuning stage to learn without forgetting, which is similar to our approach. But our goal is to generalize the model to \emph{unknown} domains and we add semantic constraints to refine the original model's response.
\vspace{-0.4cm}
\paragraph{Knowledge Distillation.}
\cite{hinton2015distilling,romero2014fitnets} first proposed to compress knowledge from a large teacher network to a small student network. Later, knowledge distillation was extended to optimizing deep networks in many generations~\cite{furlanello2018born, yang2018knowledge} and~\cite{bagherinezhad2018label} pointed out that knowledge distillation could refine ground truth labels. In ZS-SBIR, to preserve the knowledge learned at the pre-training stage, we propose to generate pseudo ImageNet labels for the training samples in the fine-tuning dataset.

%------------------------------------------------------------------------
\section{The Proposed Approach}
\label{sec:approach}

\renewcommand{\figurewidth}{16cm}
\begin{figure*}[t!]
\centering
\includegraphics[width=\figurewidth]{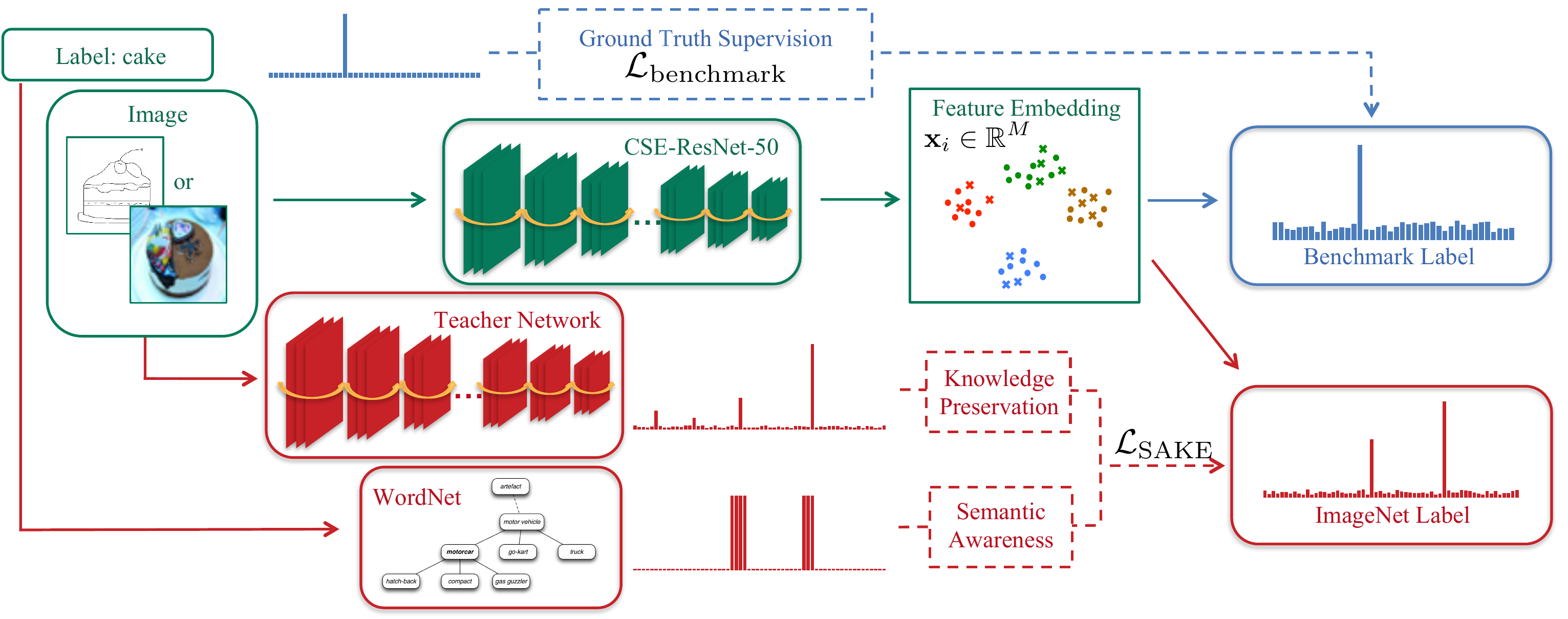}

\vspace{-0.2cm} \caption{An overview of our model. We use a CSE-ResNet-50 to embed both sketch and photo images into a shared embedding space. After obtaining the feature representation $\mathbf{x}_i$, we use it in two classification tasks, one is to predict a distribution over the benchmark labels, the other is to predict a distribution over the Imagenet labels. The former task is supervised by the ground truth in the benchmark. The latter is trained using teacher signal from an ImageNet pre-trained model and constrained by semantic information.}
\vspace{-0.2cm} \label{Fig:model}
\end{figure*}

In this section, we start with describing the problem of zero-shot sketch-based image retrieval (ZS-SBIR), and then we elaborate our motivation, which lies in the connections between zero-shot learning and catastrophic forgetting. Based on this observation, we present our solution which aims at maximally preserving knowledge from the pre-trained model, and we assist this process with weak semantic correspondence.

\subsection{Problem Statement}

In zero-shot sketch-based image retrieval (ZS-SBIR), the dataset is composed of two subsets, namely, the \textbf{reference set} for training the retrieval model, and the \textbf{testing set} for validating its performance. The reference set represents data in the \textbf{source domain}, and we denote it as $\mathcal{O}^\mathrm{S}=\{\mathcal{P}^\mathrm{S},\mathcal{S}^\mathrm{S}\}$, where $\mathcal{P}^\mathrm{S}$ and $\mathcal{S}^\mathrm{S}$ are the subsets of photos and sketches, respectively, and the superscript $\mathrm{S}$ indicates source. Similarly, the testing set contains data in the \textbf{target domain} and is denoted as $\mathcal{O}^\mathrm{T}=\{\mathcal{P}^\mathrm{T},\mathcal{S}^\mathrm{T}\}$ where the superscript $\mathrm{T}$ is for target.

During the training stage of ZS-SBIR, the photos and sketches in the reference set are used for two purposes: (i) providing semantic categories for the model to learn; and more importantly, (ii) guiding the model to realize the cross-modal similarity between photos and sketches. Mathematically, let $\mathcal{P}^\mathrm{S}=\{(\mathbf{p}_i,y_i)|y_i\in\mathcal{C}^\mathrm{S}\}^{n_1}_{i=1}$ and $\mathcal{S}^\mathrm{S}=\{(\mathbf{s}_j,z_j)|z_j\in\mathcal{C}^\mathrm{S}\}^{n_2}_{j=1}$ ($y_i$ and $z_j$ can also be written into vector form $\mathbf{y}_i=\mathbf{1}_{y_i}\in\mathbb{R}^{|\mathcal{C}^\mathrm{S}|}$, $\mathbf{z}_j=\mathbf{1}_{z_j}\in\mathbb{R}^{|\mathcal{C}^\mathrm{S}|}$), where $\mathcal{C}^\mathrm{S}$ is the reference class set. Most existing approaches trained a mapping function on these two datasets, being aware of whether the input is a photo or a sketch\footnote{This is important to improve feature extraction in the testing set. Typically there are two types of methods, {\em i.e.}, either training two networks with individual weights~\cite{dutta2019semantically, kiran2018zero} or designing internal structures in the same network for discrimination~\cite{lu2018learning}.}. During testing, a sketch query $\mathbf{s}_j'\in\mathcal{S}^\mathrm{T}$ is given at each time, and the goal is to search for the images with the same semantic label in $\mathcal{P}^\mathrm{T}$, {\em i.e.}, all $\mathbf{p}_i'\in\mathcal{P}^\mathrm{T}$ so that $y_i'=z_j'$, where both $y_i'$ and $z_j'$ fall within the testing class set $\mathcal{C}^\mathrm{T}$. The zero-shot setting indicates that no testing class appears in the training stage, {\em i.e.}, $\mathcal{C}^\mathrm{S}\cap\mathcal{C}^\mathrm{T}=\varnothing$.

\iffalse

\begin{align*}
    \mathbf{w}^T\mathbf{x}_i+\mathbf{b} = \mathbf{\hat{r}}_i,~~~~~~
    \mathbf{v}^T\mathbf{x}_i+\mathbf{a} = \mathbf{\hat{q}}_i
\end{align*}

\textcolor{red}{$\mathbf{\hat{r}}_i$ is a vector of length $|\mathcal{C}^c|$, $\mathbf{\hat{q}}_i$ is a vector of length $|\mathcal{C}^\text{IN}|$. They are the logits before softmax layer which indicate the probability of $p(y_i)=k$ for $k\in \mathcal{C}^c$, or $p(y_i)=m$ for $m\in \mathcal{C}^\text{IN}$, respectively.  Considering the semantic correlation between benchmark category $k$ and ImageNet category $m$, a simple linear mapping can be written down as $p(y_i=k) = \sum_{m}\alpha_{mk} p(y_i=m) + \beta_k$, which implies the correlation between ($\mathbf{w}$, $\mathbf{b}$) and ($\mathbf{v}$, $\mathbf{a}$).}
\fi

\subsection{Motivation: the Connection between Zero-shot Learning and Catastrophic Forgetting}

We aim at learning two models, denoted by $\mathbf{f}(\cdot;\boldsymbol{\theta}_\mathrm{P})$ and $\mathbf{g}(\cdot;\boldsymbol{\theta}_\mathrm{S})$, for feature extraction from photo and sketch images, respectively. We assume both $\mathbf{f}(\cdot;\boldsymbol{\theta}_\mathrm{P})$ and $\mathbf{g}(\cdot;\boldsymbol{\theta}_\mathrm{S})$ are deep networks which output vectors of the same dimensionality, $M$. This is to say, each learned feature vector, either $\mathbf{x}_i=\mathbf{f}(\mathbf{p}_i;\boldsymbol{\theta}_\mathrm{P})$ or $\mathbf{x}_j=\mathbf{g}(\mathbf{s}_j;\boldsymbol{\theta}_\mathrm{S})$, is an element in $\mathbb{R}^M$.

We note that during testing, the distance between these features is computed to measure the similarity between the sketch query and each photo candidate. This is to say, the goal of SBIR is to train the feature extractors so that features of the same class are projected close to each other in $\mathbb{R}^M$. With a reference set available, training a classification model is a straightforward solution. However, due to the limited amount of training data in the source set, researchers often borrow a pre-trained model from ImageNet \cite{deng2009imagenet}, a large-scale image database, and fine-tune the model to the source domain. Consequently, after training, we obtain a model capable of feature representation in the source domain, but this does not necessarily imply satisfying performance in the target domain, particularly in the zero-shot setting that these two domains rarely overlap.

From the analysis above, it is clear that our goal is to bridge the gap between the seen source domain and the unseen target domain. As the latter remains invisible, we turn to observe the behavior of domain adaptation in another visible domain. The natural choice lies in the original domain of ImageNet. We ask that after being fine-tuned on the source domain, how good the model is at representing the original domain. However, the fine-tuned model reports unsatisfying performance in the original domain, even provided that the pre-trained model was trained by the same data. This phenomenon was named catastrophic forgetting~\cite{kirkpatrick2017overcoming,french1999catastrophic}, which claims that previously acquired knowledge is mostly eliminated after the model is tuned to another domain. To verify this, as stated in the Introduction, we train an AlexNet~\cite{krizhevsky2012imagenet} on ImageNet and then fine-tune it on the TU-Berlin~\cite{eitz2012hdhso} reference set. Then, we extract the {\tt fc7} features and train a linear classifier for ImageNet on top of these fixed features. The dramatic drop of classification accuracy (from $56.29\%$ to $45.54\%$) verifies that a part of knowledge learned from ImageNet is forgotten (\emph{i.e.}, not preserved).

This motivates us to conjecture that zero-shot learning is closely related to catastrophic forgetting. In other words, by alleviating catastrophic forgetting, the ability to adapt back to the original domain becomes stronger, thus we can also expect the ability to transfer to the target domain becomes stronger. Note, to honor the zero-shot setting, the category set in the original domain and the category set in the target domain are also required to be exclusive, \emph{i.e.}, $\mathcal{C}^\mathrm{O}\cap\mathcal{C}^\mathrm{T} = \varnothing$, where the superscript O stands for original. We implement this idea using a simple yet effective algorithm, which is detailed in the next subsection.

%------------------------------------------------------------------------

\subsection{Semantic-Aware Knowledge Preservation}
We first describe the network architecture we use for feature extraction.
Most previous works in SBIR~\cite{zhang2016sketchnet, sketchy2016, liu2017deep, zhang2018generative, shen2018zero} use independent networks or semi-heterogeneous networks (networks that have independent lower levels and aggregate at top levels) to process the photos and sketches separately. Here, we adopt the Conditional SE (CSE) module proposed in \cite{lu2018learning} and integrate it into ResNet blocks to get a simple CSE-ResNet-50 network, which is used to process the photos and sketches jointly. CSE utilizes two fully connected layers, followed by a sigmoid activation to re-weight the importance of channels after each block. During the forward pass, a binary code is appended to the output of the first layer to indicate the domain of the input data, i.e., whether it is a photo or a sketch. Thus, instead of having two independent networks $\mathbf{f}(\cdot;\boldsymbol{\theta}_\mathrm{P})$ and $\mathbf{g}(\cdot;\boldsymbol{\theta}_\mathrm{S})$, what we have is a unified network  $\mathbf{h}(\cdot,\cdot;\boldsymbol{\theta})$ by letting $\mathbf{f}(\cdot;\boldsymbol{\theta}_\mathrm{P}) = \mathbf{h}(\cdot, \text{input\_domain}=0;\boldsymbol{\theta})$ and $\mathbf{g}(\cdot;\boldsymbol{\theta}_\mathrm{S}) = \mathbf{h}(\cdot, \text{input\_domain}=1;\boldsymbol{\theta})$. This conditional auto-encoder structure helps the network to learn different characteristics in input data coming from different modalities. Experiments in~\cite{lu2018learning} verified the effectiveness of CSE.

After obtaining the feature representation $\mathbf{x}_i = \mathbf{h}(\mathbf{p}_i,0;\boldsymbol{\theta})$ (or $ \mathbf{h}(\mathbf{s}_i,1;\boldsymbol{\theta})$ for sketch input) using the CSE-ResNet-50, the network forks into two classifiers: one is to predict the benchmark label $y_i \text{ (or }z_i\text{)}\in \mathcal{C}^\mathrm{S}$ for the photo $\mathbf{p}_i$ (or sketch $\mathbf{s}_i$); the other is to predict the ImageNet label, \emph{i.e.}, how likely the data belongs to each of the $1000$ ImageNet classes $\mathcal{C}^\mathrm{O}$. Both branches are constructed by adding one fully connected layer on top of $\mathbf{x}_i$, followed by a softmax function. More specifically, the first classifier $\mathcal{W}^\mathrm{B}$ computes $\hat{\mathbf{y}}_i=\text{softmax}(\boldsymbol{\alpha}^\top\mathbf{x}_i+\boldsymbol{\beta})$, $\hat{\mathbf{y}}_i\in\mathbb{R}^{|\mathcal{C}^\mathrm{S}|}$, and aims to adapt the network to the SBIR benchmarks, especially the reference set, achieving the goal of bridging the gap between sketch and photo images and learning good similarity measure for cross-modality data. The second classifier $\mathcal{W}^\mathrm{I}$ works on $\Tilde{\mathbf{y}}_i = \text{softmax}(\boldsymbol{\zeta}^\top\mathbf{x}_i+\boldsymbol{\eta})$, $\Tilde{\mathbf{y}}_i\in\mathbb{R}^{|\mathcal{C}^\mathrm{O}|}$, which helps to preserve the network's capability of recognizing the rich visual features learned from previous ImageNet training, benefiting the network's adaptation to ZS-SBIR target domain. $\boldsymbol{\alpha}, \boldsymbol{\beta}, \boldsymbol{\zeta}, \boldsymbol{\eta}$ are weights and bias terms in the two linear classifiers, respectively.

Without access to the original ImageNet data, we argue the training of the second classifier is non-trivial. To solve the problem of having no ground truth ImageNet label for images in the benchmark dataset, SAKE inquires an ImageNet pre-trained model, \emph{i.e.}, the model SAKE is initialized from, to provide teacher signal, which, after refined by semantic constraints, is used to supervise the learning of $\Tilde{\mathbf{y}}$. Next, we explain the training objective in detail.

\subsection{Objective of Optimization}

The two classification tasks are trained end-to-end simultaneously, and the learning objective of our model can be written as $\mathcal{L} = \mathcal{L}_\mathrm{benchmark} + \lambda_\mathrm{SAKE}\mathcal{L}_\mathrm{SAKE}$, where $\mathcal{L}_\mathrm{benchmark}$ models the classification loss in $\hat{\mathbf{y}}$ based on the ground-truth. We compute it using the cross-entropy loss function:
\begin{align*}
    \mathcal{L}_\mathrm{benchmark} = \frac{1}{N}\sum_i-\log\frac{\exp{(\boldsymbol{\alpha}^\top_{y_i}\mathbf{x}_i+\beta_{y_i})}}{\sum_{k\in\mathcal{C}^\mathrm{S}}\exp{(\boldsymbol{\alpha}^\top_k\mathbf{x}_i+\beta_k)}},
\end{align*}
where $N$ is the total training sample number, $\boldsymbol{\alpha}_k$ and $\beta_k$ are the weight and bias terms in the benchmark label classifier $\mathcal{W}^\mathrm{B}$ for category $k$. $y_i$ can be replaced by $z_i$ if the input data is a sketch.

$\mathcal{L}_\mathrm{SAKE}$ computes the classification loss in $\Tilde{\mathbf{y}}$. Since no ground truth label is available for this loss term, we combine a teacher signal and semantic constraints into it. In what follows, we elaborate these two components in details.

\vspace{-0.4cm}
\paragraph{Learning from a Teacher Signal.}
Given a photo image with unknown object label among the $1000$ ImageNet classes $\mathcal{C}^\mathrm{O}$, it is intuitive to use an ImageNet trained classifier to estimate its identity. Inspired by the recent work in \emph{knowledge distillation}~\cite{furlanello2018born,hinton2015distilling} and \emph{incremental learning}~\cite{li2018learning, shmelkov2017incremental}, we propose to achieve this goal by using the ImageNet pre-trained network as a teacher, \emph{i.e.}, teach our model to remember the rich visual features and make reasonable ImageNet label predictions. During the training process, the teacher network is fixed and takes the same photo input as the model does. According to the prediction $\mathbf{q}^\mathrm{t}_i = \mathrm{Softmax}(\mathbf{t}_{i})\in\mathbb{R}^{|\mathcal{C}^\mathrm{O}|}$ made by the teacher network, {\em i.e.}, the probability of sample $\mathbf{p}_i$ belongs to each category in $\mathcal{C}^{\mathrm{O}}$, we encourage our model to make the same prediction. Unlike ground truth labels which are one-hot vectors, what we get from the teacher network is a discrete probability distribution over $\mathcal{C}^\mathrm{O}$. Therefore, the cross-entropy loss with soft labels is used to compute the teacher loss:
\begin{align*}
    \mathcal{L}_\mathrm{teacher}=\frac{1}{N}\sum_{i}\sum_{m\in\mathcal{C}^\mathrm{O}}-q_{i,m}^\mathrm{t}\log\frac{\exp{(\boldsymbol{\zeta}^\top_m\mathbf{x}_i+\eta_m)}}{\sum_{l\in\mathcal{C}^\mathrm{O}}\exp{(\boldsymbol{\zeta}^\top_l\mathbf{x}_i+\eta_l)}},
\end{align*}
where $\boldsymbol{\zeta}_m$ and $\eta_m$ are the weight and bias terms in the ImageNet label classifier $\mathcal{W}^\mathrm{I}$ for category $m$. Since random transformation is added to each input sample for data augmentation purpose, the teacher network makes predictions online. During the test step, no teacher network is needed.

\vspace{-0.4cm}
\paragraph{Semantic Constraints of the Teacher Signal.}
Although the teacher network has been trained on the sophisticated ImageNet dataset, there is a knowledge gap between the original domain and the source domain, so it may make mistakes on the SBIR reference set. The supervision given by the wrong predictions made by the teacher will hurt the goal of preserving useful original domain knowledge in our SAKE model. Therefore, we propose to use additional semantic information to guide the teacher-student optimization process. More specifically, we use WordNet~\cite{miller1995wordnet, miller1998wordnet} to construct a semantic similarity matrix $\mathbf{A}$; each entry $a_{k,m}$ represents the similarity between class $k\in\mathcal{C}^\mathrm{S}$ and class $m\in\mathcal{C}^\mathrm{O}$. Given a benchmark sample $\mathbf{p}_i$ with ground truth label $y_i=k$, we encourage the prediction of $\Tilde{y}_{im}$ to be large if class $m$ is semantically similar to $k$, \emph{i.e.},~$a_{k,m}$ is large. 

$\mathbf{a}_{k}$ is defined for each class and can be combined with $\mathbf{t}_{i}$ to form the semantic-aware teacher signal where the logits is a weighted sum of the two components, $\mathbf{q}_{i}=\mathrm{Softmax}(\lambda_1\cdot\mathbf{t}_i + \lambda_2\cdot\mathbf{a}_{y_i})$. Therefore, the SAKE loss can be written down as:
\begin{align*}
    \mathcal{L}_\mathrm{SAKE}=\frac{1}{N}\sum_{i}\sum_{m\in\mathcal{C}^\mathrm{O}}-q_{i,m}\log\frac{\exp{(\boldsymbol{\zeta}^\top_m\mathbf{x}_i+\eta_m)}}{\sum_{l\in\mathcal{C}^\mathrm{O}}\exp{(\boldsymbol{\zeta}^\top_l\mathbf{x}_i+\eta_l)}},
\end{align*}
where $\boldsymbol{\zeta}_m$ and $\eta_m$ are the same as defined in the teacher loss, are the weight and bias terms in the ImageNet label classifier for category $m$. Note $\mathcal{L}_\mathrm{teacher}$ is a special setting of $\mathcal{L}_\mathrm{SAKE}$ with $\lambda_1=1$ and $\lambda_2=0$. We argue this loss term helps to refine the supervision signal from the teacher network and makes the knowledge preservation process semantic aware.

%------------------------------------------------------------------------

%------------------------------------------------------------------------

% \subsection{Optimization}

%------------------------------------------------------------------------

\newcommand{\colwidthA}{1.1cm}
\begin{table*}[]
\centering{
\setlength{\tabcolsep}{0.12cm}
\begin{tabular}{lccccccccc}
\hline
\multirow{2}{*}{Method} & \multirow{2}{*}{SBIR} & \multirow{2}{*}{Zero-Shot} & \multirow{2}{*}{Dimension} & \multicolumn{2}{c}{TU-Berlin Ext.} & \multicolumn{2}{c}{Sketchy Ext.} & \multicolumn{2}{c}{Sketchy Ext. (\cite{kiran2018zero} Split)} \\ \cline{5-10} 
                        &                       &                            &                            & mAP@all      & Prec@100      & mAP@all     & Prec@100     & mAP@200             & Prec@200             \\ \hline
GN-Triplet~\cite{sketchy2016}              & Yes                   & No                         & $1024$                       & $0.189$        & $0.241$              & $0.211$       & $0.310$              & $0.083$               & $0.169$                     \\
DSH~\cite{liu2017deep}                     & Yes                   & No                         & $64\dagger$               & $0.122$        & $0.198$              & $0.164$       & $0.227$             & $0.059$               & $0.153$                     \\
SAE~\cite{kodirov2017semantic}                     & No                    & Yes                        & $300$                        & $0.161$        & $0.210$               & $0.210$        & $0.302$             & $0.136$               & $0.238$                     \\
ZSH~\cite{yang2016zero}                     & No                    & Yes                        & $64\dagger$             & $0.139$        & $0.174$              & $0.165$       & $0.217$             & -                   & -                         \\
ZSIH~\cite{shen2018zero}                    & Yes                   & Yes                        & $64\dagger$                 & $0.220$         & $0.291$              & $0.254$       & $0.340$              & -                   & -                         \\
\multirow{2}{*}{EMS~\cite{lu2018learning}} & \multirow{2}{*}{Yes} & \multirow{2}{*}{Yes} & $512$                        & $0.259$        & $0.369$              & -           & -                 & -                   & -                         \\
                        &                       &                            & $64\dagger$             & $0.165$        & $0.252$              & -           & -                 & -                   & -                         \\
CAAE~\cite{kiran2018zero}                    & Yes                   & Yes                        & $4096$                       & -            & -                  & $0.196$           & $0.284$                 & $0.156$               & $0.260$                      \\
CVAE~\cite{kiran2018zero}                    & Yes                   & Yes                        & $4096$                       & -            & -                  &     -       & -                 & $0.225$               & $0.333$                     \\
\multirow{2}{*}{SEM-PCYC~\cite{dutta2019semantically}} & \multirow{2}{*}{Yes} & \multirow{2}{*}{Yes} & $64$                       & $0.297$            & $0.426$                  & $0.349$           & $0.463$                 & -               & -                      \\
                    &                    &                         & $64\dagger$                       & $0.293$            & $0.392$                  & $0.344$           & $0.399$                 & -               & -                     \\\hline
\multirow{2}{*}{\textbf{SAKE}} & \multirow{2}{*}{Yes} & \multirow{2}{*}{Yes} & $512$                        &      $\mathbf{0.475}$        &    $\mathbf{0.599}$                &       $\mathbf{0.547}$      &         $\mathbf{0.692}$          &           $\mathbf{0.497}$          &            $\mathbf{0.598}$               \\
                        &                       &                            & $64\dagger$                  &       $\mathbf{0.359}$       &         $\mathbf{0.481}$           &        $\mathbf{0.364}$     &         $\mathbf{0.487}$          &          $\mathbf{0.356}$           &               $\mathbf{0.477}$            \\\hline
\end{tabular}
}
\vspace{-0.2cm} \caption{ZS-SBIR performance comparison of SAKE and existing methods. ``$\dagger$'' denotes experiments using binary hashing codes. The rest use real-valued features. ``-'' indicates the results are not presented by the authors on that metric.}
\vspace{-0.2cm} \label{tab:compare1}
\end{table*}

\section{Experiments}
\label{sec:experiments}
\subsection{Datasets and Settings}

\paragraph{Datasets.} We evaluated SAKE on two large-scale sketch-photo datasets: TU-Berlin~\cite{eitz2012hdhso} and Sketchy~\cite{sketchy2016} with extended images obtained from \cite{zhang2016sketchnet, liu2017deep}. The TU-Berlin dataset contains 20,000 sketches uniformly distributed over 250 categories. The additional 204,489 photo images provided in \cite{zhang2016sketchnet} are also used in our work. The Sketchy dataset consists of 75,471 hand-drawn sketches and 12,500 corresponding photo images from 125 categories. Additional 60,502 photo images were collected by~\cite{liu2017deep}, yielding a total of 73,002 samples. For comparison, we follow \cite{shen2018zero} and randomly pick 30/25 classes as the testing set from TU-Berlin/Sketchy, and use the rest 220/100 classes as the reference set for training. During the testing step, the sketches from the testing set are used as the retrieval queries, and photo images from the same set of classes are used as the retrieval gallery. As \cite{shen2018zero} suggested, each class in the testing set is required to have at least 400 photo images.

It comes to our attention that some categories in the TU-Berlin/Sketchy are also present in the ImageNet dataset, and if we select them as our testing set, it will violate the zero-shot assumption (the same for the existing works that use an ImageNet pre-trained model for initialization). Thus, we follow the work of \cite{kiran2018zero} and test our model using their careful split in Sketchy, which includes $21$ testing classes that are not present in the $1000$ classes of ImageNet. The performance of SAKE on this careful split of Sketchy is shown in the result section. For the TU-Berlin dataset, we also carefully evaluate the performance of our model when applied to a testing set that is composed of ImageNet and non-ImageNet classes. The results can be found in Section \ref{sec:quantitative}.

\vspace{-0.4cm}
\paragraph{Implementation Details.}
We implemented our model using PyTorch \cite{paszke2017automatic} with two TITAN X GPUs. We use a SE-ResNet-50 network pretrained on ImageNet to initialize our model, and it is also used as the teacher network in SAKE during the training stage. To provide the semantic constraints, \texttt{WordNet} python interface from \texttt{nltk} corpus reader is used to measure the similarity between object category labels. We map each category to a node in WordNet and use the {\sf path\_similarity} to set $a_{k,m}$. To train our model, Adam optimizer is applied with parameters $\beta_1 = 0.9$, $\beta_2 = 0.999$, $\lambda = 0.0005$. The learning rate starts at $0.0001$ and exponentially decayed to $1\mathrm{e}-7$ during training. We use batch size equals $40$ and train networks for $20$ epochs. In our experiments, $\lambda_\mathrm{SAKE}$ is set to $1$, $\lambda_1$ is set to $1$, $\lambda_2$ is set to $0.3$, unless stated otherwise.

To achieve ZS-SBIR, nearest neighbor search is conducted based on distance calculated by $\mathbf{x}_i$. For real-valued feature vectors, cosine distance is used to avoid variations introduced by the vector norm. To accelerate the retrieval speed, binary hashing is widely used to encode the input data. To make fair comparisons to existing zero-shot hashing methods \cite{shen2018zero, yang2016zero}, we apply the iterative quantization (ITQ) \cite{gong2013iterative} algorithm on the feature vectors learned by our model to obtain the binary codes. Following \cite{dutta2019semantically}, we use the final representations of sketches and photo from the training set to learn an optimized rotation, which is then used on the feature vectors of testing samples to obtain the binary codes. After that, hamming distance is calculated for the retrieval task. We will release our models and codes upon acceptance.

%------------------------------------------------------------------------

\subsection{Comparison with Existing Methods}
We compare our model with three prior works on ZS-SBIR: ZSIH~\cite{shen2018zero}, CAAE and CVAE~\cite{kiran2018zero}, and SEM-PCYC~\cite{dutta2019semantically}, which all use generative models and complicated frameworks, \emph{e.g.}, graph conv layers, adversarial training, etc., to encourage the learning of good shared embedding space. EMS proposed by \cite{lu2018learning} is the current state-of-the-art model in SBIR and is claimed to be able to address zero-shot problems directly, so we include their ZS-SBIR results for comparison. We also compare our model with two SBIR methods, GN-Triplet~\cite{sketchy2016} and DSH~\cite{liu2017deep}, and two zero-shot methods, SAE~\cite{kodirov2017semantic} and ZSH~\cite{yang2016zero}. All models use ImageNet pre-trained network for weights initialization. Mean average precision (mAP@all) and precision considering top 100 retrievals (Precision@100) are computed for performance evaluation and comparison.

As results shown in Table \ref{tab:compare1}, despite the simple design of our framework, in all datasets/dataset splits, our proposed method consistently outperforms the state-of-the-art ZS-SBIR methods by a large margin, \emph{e.g.}, $20.9\%$ relative improvement of mAP@all for the challenging TU-Berlin Extension dataset using $64$-bit binary hashing codes. To address the concern that most works use their own random reference/testing split without publishing the experimental details, we repeat our experiment on TU-Berlin Extension three times using different random splits, and get mAP@all equals $0.352$, $0.369$, $0.359$, in the $64$-bit binary case, which all outperform the previous models. This confirms the large performance gain of our SAKE model is not by chance or by split bias.

\renewcommand{\colwidthA}{1.75cm}
\begin{table}[]
\centering{
\setlength{\tabcolsep}{0.14cm}
\begin{tabular}{lcccccc}
\hline
     & \multicolumn{3}{c}{TU-Berlin Ext.} & \multicolumn{3}{c}{Sketchy Ext.} \\ \cline{2-7} 
     & $32$   & $64$   & $128$  & $32$  & $64$  & $128$  \\ \hline
ZSH~\cite{yang2016zero}  & $0.132$     & $0.139$     & $0.153$     & $0.146$    & $0.165$    & $0.168$     \\
ZSIH~\cite{shen2018zero} & $0.201$     & $0.220$     & $0.234$     & $0.232$    & $0.254$    & $0.259$     \\ \hline
\textbf{SAKE} & $0.269$     & $0.359$     & $0.392$     & $0.289$    & $0.364$    & $0.410$     \\ \hline
\end{tabular}
}
\vspace{-0.2cm} \caption{ZS-SBIR mAP@all comparison of SAKE and existing zero-shot hashing methods. $32$, $64$, and $128$ represent the length of the generated hashing codes.}
\vspace{-0.2cm} \label{tab:compare2}
\end{table}

\begin{table}[]
\small
\centering{
\setlength{\tabcolsep}{0.12cm}
\begin{tabular}{lcccccc}
\hline
              & $\lambda_1$            & \multicolumn{5}{c}{$\lambda_2$} \\ \cline{3-7} 
              &            & $0$       & $0.1$     & $0.3$     & $1$      & $3$      \\ \hline
ZS-SBIR   & $0$ & $0.362$   & $0.364$ & $0.370$ & $0.369$ & $0.362$  \\ \hline
ZS-SBIR   & $1$ & $0.426$   & $0.431$   & $\mathbf{0.434}$    & $0.416$  & $0.412$  \\ \hline
\end{tabular}
\caption{ZS-SBIR mAP@all on TU-Berlin Extension dataset with different $\lambda_1$ and $\lambda_2$. $\lambda_\mathrm{SAKE}=1$ for all tests.}
\label{tab:lambda_2}
\vspace{-0.4cm}
}
\end{table}

\renewcommand{\colwidthA}{1.5cm}
\newcommand{\colwidthB}{2.3cm}
\begin{table*}[]
\centering{
\setlength{\tabcolsep}{0.08cm}
\begin{tabular}{lC{\colwidthA}C{\colwidthA}C{\colwidthA}C{\colwidthB}|C{\colwidthA}C{\colwidthA}C{\colwidthA}C{\colwidthB}}
\hline
\multirow{2}{*}{BackBone} & \multicolumn{4}{c|}{ZS-SBIR}               & \multicolumn{4}{c}{ZS-PBIR}               \\ \cline{2-9} 
                          & pretrained & $\mathcal{L}_\mathrm{B}$  & $\mathcal{L}_\mathrm{B}+\mathcal{L}_\mathrm{T}$ & $\mathcal{L}_\mathrm{B}+\mathcal{L}_\mathrm{SAKE}$ & pretrained & $\mathcal{L}_\mathrm{B}$  & $\mathcal{L}_\mathrm{B}+\mathcal{L}_\mathrm{T}$ & $\mathcal{L}_\mathrm{B}+\mathcal{L}_\mathrm{SAKE}$ \\ \hline \hline
AlexNet                   &      $0.074$      & $0.267$ &     $0.275$      &     $0.275$     &     $0.386$        & $0.393$ &       $0.427$     &        $0.432$   \\
ResNet-50                  &     $0.081$       & $0.352$ &     $0.395$      &     $0.413$     &         $0.640$    & $0.542$ &      $0.666$      &        $0.670$   \\
CSE-ResNet-50              &     $0.068$       & $0.353$ & $0.426$     &     $\mathbf{0.434}$     &      $0.635$       & $0.558$ & $0.673$     &       $\mathbf{0.683}$    \\ \hline
\end{tabular}
}
\vspace{-0.2cm} \caption{ZS-SBIR and ZS-PBIR mAP@all on TU-Berlin Extension for different backbone models with different loss terms. All models are pre-trained using ImageNet, and represent each image by a $64$-d feature vector. $\mathcal{L}_\mathrm{B}$ stands for $\mathcal{L}_\mathrm{benchmark}$. $\mathcal{L}_\mathrm{T}$ stands for $\mathcal{L}_\mathrm{teacher}$.}
\vspace{-0.2cm} \label{tab:loss}
\end{table*}

Since the categories in both TU-Berlin and Sketchy overlap with ImageNet, it is important to test the model using non-ImageNet categories as testing to honor the zero-shot assumption, especially for our SAKE model which largely relies on knowledge from the original domain, \emph{i.e.}, rich visual features learned previously from ImageNet. Thus, in Table \ref{tab:compare1}, we reported our model's performance on this careful split proposed by \cite{kiran2018zero}, which only use classes that are not present in the $1000$ classes of ImageNet as testing. The result shows SAKE still outperforms the baselines by a large margin. This result demonstrates that the original domain knowledge preserved by SAKE is not only maintaining its ability to be adapted back to the original domain but also helping the model to be more generalizable to the \emph{unseen} target domain.

In Table \ref{tab:compare2}, we further compare our model with the two zero-shot hashing methods, ZSH\cite{yang2016zero} and ZSIH~\cite{shen2018zero}, using binary codes of different lengths. As expected, longer hashing code leads to better retrieval performance, and our proposed model beats both methods in all cases. This again proves the effectiveness of the proposed SAKE model. Diagnostic experiments are given in the following subsections to show the superior performance of SAKE is indeed from the knowledge preserved with semantic constraints.

%------------------------------------------------------------------------
\renewcommand{\colwidthA}{1.8cm}
\subsection{Quantitative Analysis}
\label{sec:quantitative}
\begin{table}[]
\centering{
\setlength{\tabcolsep}{0.08cm}
\begin{tabular}{lccccc}
\hline
                          & \multicolumn{5}{c}{$\lambda_\mathrm{SAKE}$} \\ \cline{2-6} 
                          & $0$       & $0.1$     & $0.3$     & $1$      & $3$      \\ \hline
ZS-SBIR   & $0.353$   & $0.378$   & $0.395$    & $0.434$  & $0.429$  \\
ZS-PBIR (non-IN) & $0.558$   & $0.587$   & $0.612$   & $0.654$  & $0.668$   \\
ZS-PBIR (IN)  & $0.545$   & $0.543$   & $0.615$   & $0.707$  & $0.758$  \\ \hline
\end{tabular}
\vspace{-0.2cm} \caption{mAP@all on TU-Berlin Extension with different $\lambda_\mathrm{SAKE}$. }
\vspace{-0.2cm} \label{tab:lambda}
}
\end{table}

\paragraph{Knowledge Preservation Using SAKE.} We first run a simple experiment to show the phenomenon of catastrophic forgetting during the model fine-tuning process and how SAKE helps to alleviate it. We train a linear classifier for ImageNet $1000$ classes using features extracted from the last fully connected layer of the DNNs and use the top $1$ prediction accuracy to measure the effectiveness of the features to represent the data. An ImageNet pre-trained AlexNet achieves a top-1 accuracy of $56.29\%$, while a fine-tuned model (trained to classify the $220$ object categories in TU-Berlin Extension reference set) only reports $45.54\%$. Lastly, we fine-tune the AlexNet by SAKE, and the top-1 accuracy is improved to $51.39\%$. By changing AlexNet to the deeper model SE-ResNet-50, we observe similar results: pre-trained model achieves $77.43\%$, fine-tuned model drops to $59.56\%$, and training by SAKE improves it to $67.44\%$. The result suggests benchmark training does lead to knowledge elimination for the previously learned task(s), and SAKE is able to alleviate it effectively.

\vspace{-0.4cm}
\paragraph{Ablation Study.} In Table \ref{tab:lambda_2}, we analyze the effect of hyper-parameter $\lambda_1$ and $\lambda_2$. When $\lambda_1$ is set to $0$, applying the semantic constraint without a teacher signal barely affects the results. When $\lambda_1=1$, the semantic constraint provides a mild boost with peak value at $\lambda_2=0.3$. In Table \ref{tab:loss}, we show zero-shot image retrieval mAP@all for networks with different backbones and loss terms. All networks are trained and tested in the same setting, \emph{i.e.}, using the same dataset split on TU-Berlin Extension and $64$-d feature representation. We first observe that ResNet-50 reaches better performances than AlexNet, suggesting networks with deeper architectures perform better due to their larger modeling capacities. The results we reported here for SAKE using CSE-ResNet-50 network can probably be further improved if we choose to use deeper backbones. Secondly, we find that the CSE module is effective in enhancing cross-modal mapping. It provides additional information about the data type and allows the model to learn more flexible functions to process data in each modality, so it is an important component in our SAKE design. Lastly, the results show knowledge preservation with simple unconstrained teacher signal can effectively improve the performance of all backbones, especially the one with larger capacity and higher flexibility. On top of this, semantic awareness brings in an extra boost and finally builds up our full SAKE model that reaches the best retrieval results.

\vspace{-0.4cm}
\paragraph{Why SAKE?} To further investigate how the model benefits from the original domain knowledge preserved by SAKE, we look into zero-shot photo-based image retrieval (ZS-PBIR) and use it to evaluate the representations of photo images learned by SAKE. In the ideal case, if the model is capable of recognizing the rich visual features in images over a large collection of object categories, \emph{i.e.}, the ImageNet dataset, it will apply them to the \emph{unseen} photo images and project the ones with similar visual contents into a clustered region in the embedding space. This will help the model reach good ZS-PBIR result. Indeed, as shown in Table \ref{tab:loss}, pre-trained models have reasonable mAP@all ($\boldsymbol{\zeta}$ and weight for $\mathbf{x}_i$ layer are initialized by decomposing the original weight matrix in the output layer), which is vulnerable to simple benchmark training. After adding the knowledge preservation term, either $\mathcal{L}_\mathrm{T}$ or $\mathcal{L}_\mathrm{SAKE}$, ZS-PBIR is improved by a large number. This implies the improvement of ZS-SBIR achieved by SAKE is mainly from the model's capability of generating more structured and tightly distributed feature representations for the photo images in the testing set.

In Table~\ref{tab:lambda}, we gradually increase $\lambda_\mathrm{SAKE}$, the coefficient of $\mathcal{L}_\mathrm{SAKE}$ in the total loss, and test how mAP@all for ZS-SBIR and ZS-PBIR changes. For ZS-SBIR, the performance increases and then reaches the peak value at $\lambda_\mathrm{SAKE}=1$. If we further increase $\lambda_{1}$, the performance starts to drop, probably because the model is too much affected by the teacher signal and becomes less focused on learning the new dataset. In the case of ZS-PBIR, for comparison, we specifically pick categories that are \emph{not} present in ImageNet, \emph{i.e.}, the target domain, and categories that are present in ImageNet, \emph{i.e.}, the original domain, to show how they are differently affected by $\lambda_\mathrm{SAKE}$. As we expected, the performance on ImageNet photos keeps rising as we increase $\lambda_\mathrm{SAKE}$, while the performance on non-ImageNet photos is gently boosted and then saturates faster. This result proves again that using SAKE helps to preserve the model's capability of recognizing the rich visual features in ImageNet, which is crucial for generating good representations for photo images in the \emph{unseen} retrieval gallery, resulting to largely boosted ZS-SBIR performance.

\subsection{Qualitative Analysis}
\label{subsec:diag}
\renewcommand{\figurewidth}{8cm}
\begin{figure}[!t]
\centering
\includegraphics[width=\figurewidth]{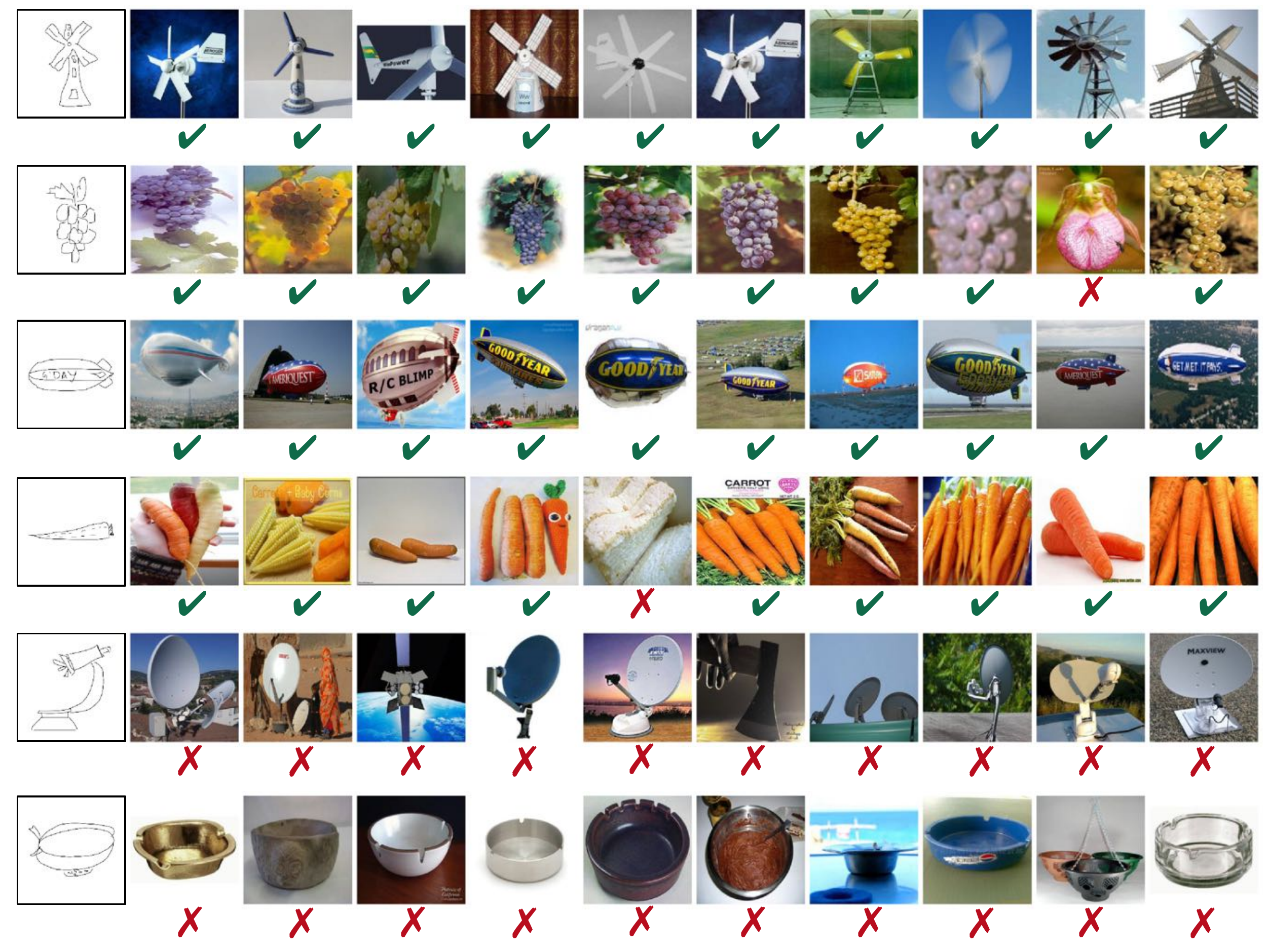}
\vspace{-0.2cm} \caption{Top-10 ZS-SBIR results obtained by SAKE on TU-Berlin Extension dataset. Retrieval is conducted by nearest neighbors search using cosine distance on $64$-d feature vectors. Green ticks denote correctly retrieved candidates, and the red crosses indicate wrong retrievals. Two negative cases are visualized here to help diagnose the model. See Section \ref{subsec:diag} for more details.}
\vspace{-0.2cm} \label{Fig:example}
\end{figure}

\renewcommand{\figurewidth}{7.6cm}
\begin{figure}[!t]
\centering
\includegraphics[width=\figurewidth]{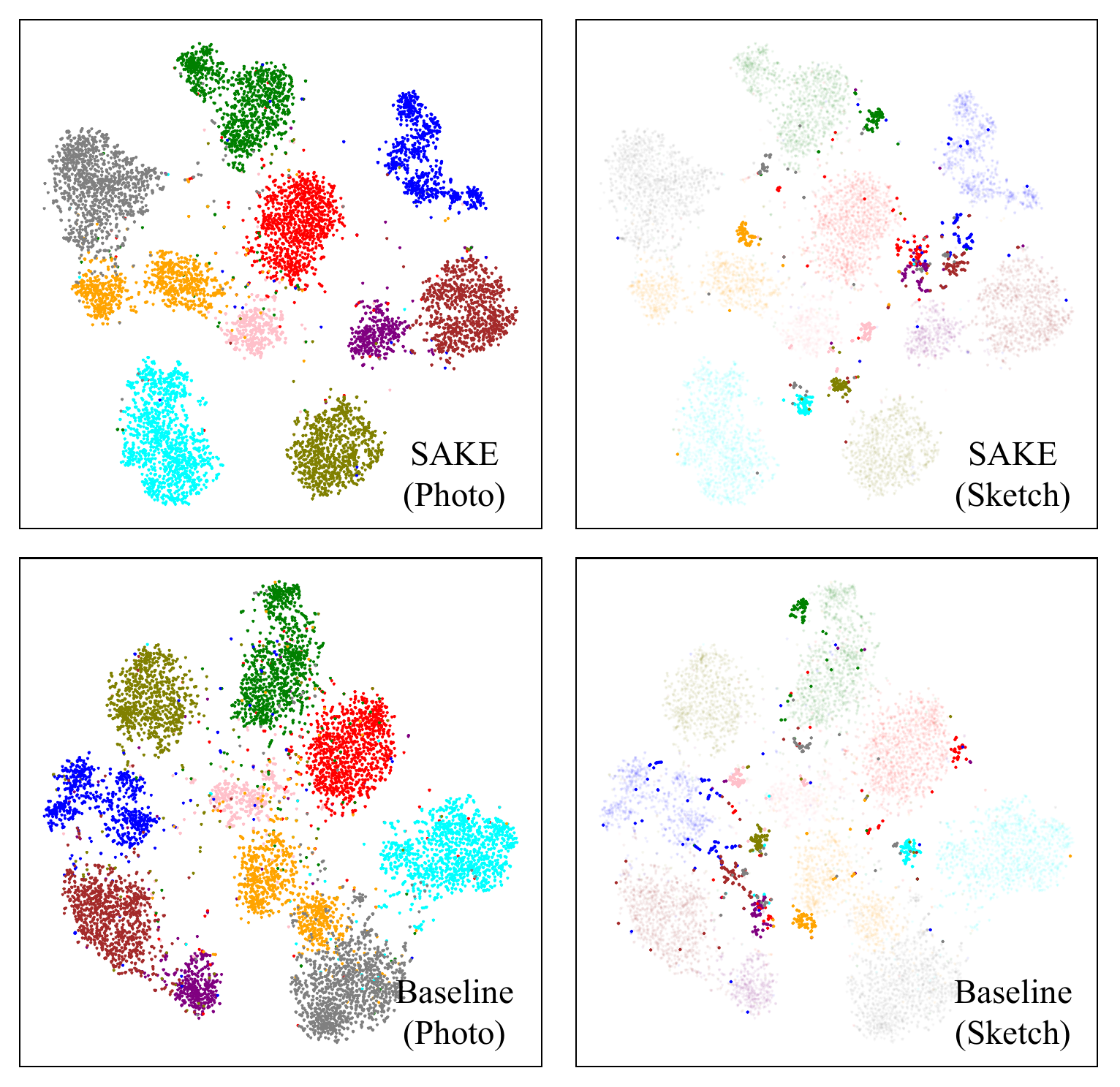}
\vspace{-0.2cm} \caption{t-SNE results using $64$-d feature representations on the testing set of TU-Berlin Extension. First row: features learned by SAKE. Second row: features learned by the baseline model without $\mathcal{L}_\mathrm{SAKE}$. In the ``Sketch'' plots, the ``Photo'' data points are retained and lightened. This figure is best viewed in color.}
\vspace{-0.2cm} \label{Fig:tsne}
\end{figure}

\paragraph{Example of Retrievals.} In Figure \ref{Fig:example}, we show the top $10$ retrieval results obtained by SAKE in the TU-Berlin Extension dataset. In most cases, SAKE retrieves photo images with the right object label, \emph{i.e.}, the same label as the sketch image has. In the selected negative cases, SAKE fails to find photo images that match the sketch category but instead returns photos from another category, which share some visual similarities with the sketch query. This implies that the feature vectors of the photos candidates are properly clustered, which benefits ZS-SBIR if sketches from the same class are also projected to the same region.

\vspace{-0.4cm}
\paragraph{Visualization of the Learned Embeddings.}
In Figure \ref{Fig:tsne}, we show the t-SNE \cite{maaten2008visualizing} results of our SAKE model compared with the baseline model using $64$-d feature representations on the testing set of TU-Berlin Extension, where a more clearly clustered map on the object classes can be found in SAKE. We also observe margins between photo and sketch data, implying SAKE could be further improved by learning more aligned features for sketches and photos.
%------------------------------------------------------------------------

\vspace{-0.2cm}
\section{Conclusions}
\label{sec:conclusions}

This paper studies the problem of zero-shot sketch-based image retrieval from a new perspective, namely, incremental learning to alleviate catastrophic forgetting. The key observation lies in the fact that both zero-shot learning and incremental learning focus on transferring the trained model to another domain, so we conjecture and empirically verify that improving the performance of the latter task benefits the former one. The proposed SAKE algorithm preserves knowledge from the original domain by making full use of semantics, so that it works without access to the original training images. Experiments on both TU-Berlin and Sketchy datasets demonstrate state-of-the-art performance. We will investigate SAKE on a wider range of tasks involving catastrophic forgetting in our future work.

One of the most important take-aways of this work is that different machine learning tasks, though look different, may reflect the same essential reason, and the reason often points to over-fitting, a long-lasting battle in learning. We shed light on a new idea, which works by dealing with one task to assist another one. We emphasize further research efforts should be devoted to this direction.

\vspace{0.2cm}
\noindent
{\bf Acknowledgement}\quad
This research was supported by NSF grant BCS-1827427. We thank Chenxi Liu for helping design Figure~\ref{Fig:Concept} and proofreading. We thank Chenglin Yang for discussions on knowledge distillation.

{\small
\bibliographystyle{ieee_fullname}
\bibliography{egbib}
}

\newpage
\onecolumn

\begin{appendices}

\section{}
This supplementary material contains extra evidences to support our claim that knowledge preservation benefits domain adaptation. It is shown in the main paper that knowledge preservation helps the fine-tuned model to maintain good performance in the original domain. Here, under the zero-shot setting, we measure the similarity between a {\bf target} zero-shot category and the {\bf original} ImageNet categories, and demonstrate how this similarity correlates to the performance improvements brought by SAKE.

\subsection{Setting}

We investigate similarities under the zero-shot setting. Different from the main paper as well as all previous work, we create a new held-out set of TU-Berlin, containing 30 categories which are {\bf not} present in ImageNet. This is to make a fair comparison between different target categories. Experiments are performed three times, {\em i.e.}, we randomly choose three held-out sets, all of which have no category overlap with ImageNet.

We take a vanilla SE-ResNet-50 model, which was pre-trained on ImageNet and fine-tuned on the TU-Berlin reference set with only $\mathcal{L}_\mathrm{benchmark}$. Then, we sort these 30 target categories by their mAP@all improvement achieved by SAKE, and equally divide them into 3 groups, ``low'', ``medium'' and ``high'', with each of them containing 10 categories of low, medium and high improvements brought by SAKE, respectively. Similarities between the target categories and the ImageNet categories are investigated separately in these 3 groups.

\setcounter{figure}{4}
\renewcommand{\figurewidth}{7cm}
\begin{figure}[!h]
\centering
\begin{subfigure}[b]{0.4\textwidth}
\includegraphics[width=\figurewidth]{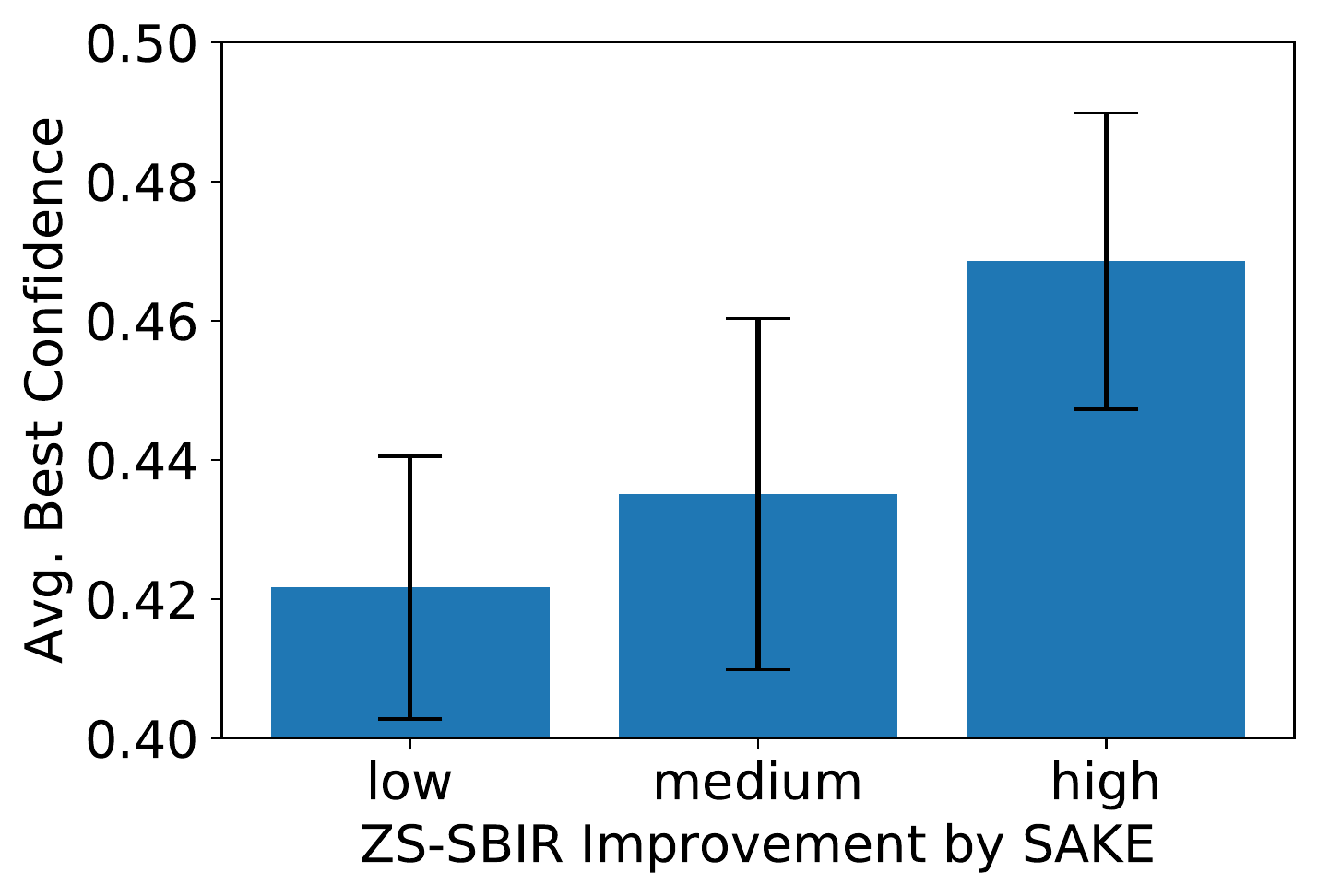}
%\vspace{-0.2cm}
\caption{ImageNet classification confidence with respect to the improvement brought by SAKE}
\label{fig:sm_visual}
\end{subfigure}
~ %add desired spacing between images, e. g. ~, \quad, \qquad, \hfill etc. 
\hspace{1cm}
\begin{subfigure}[b]{0.4\textwidth}
\includegraphics[width=\figurewidth]{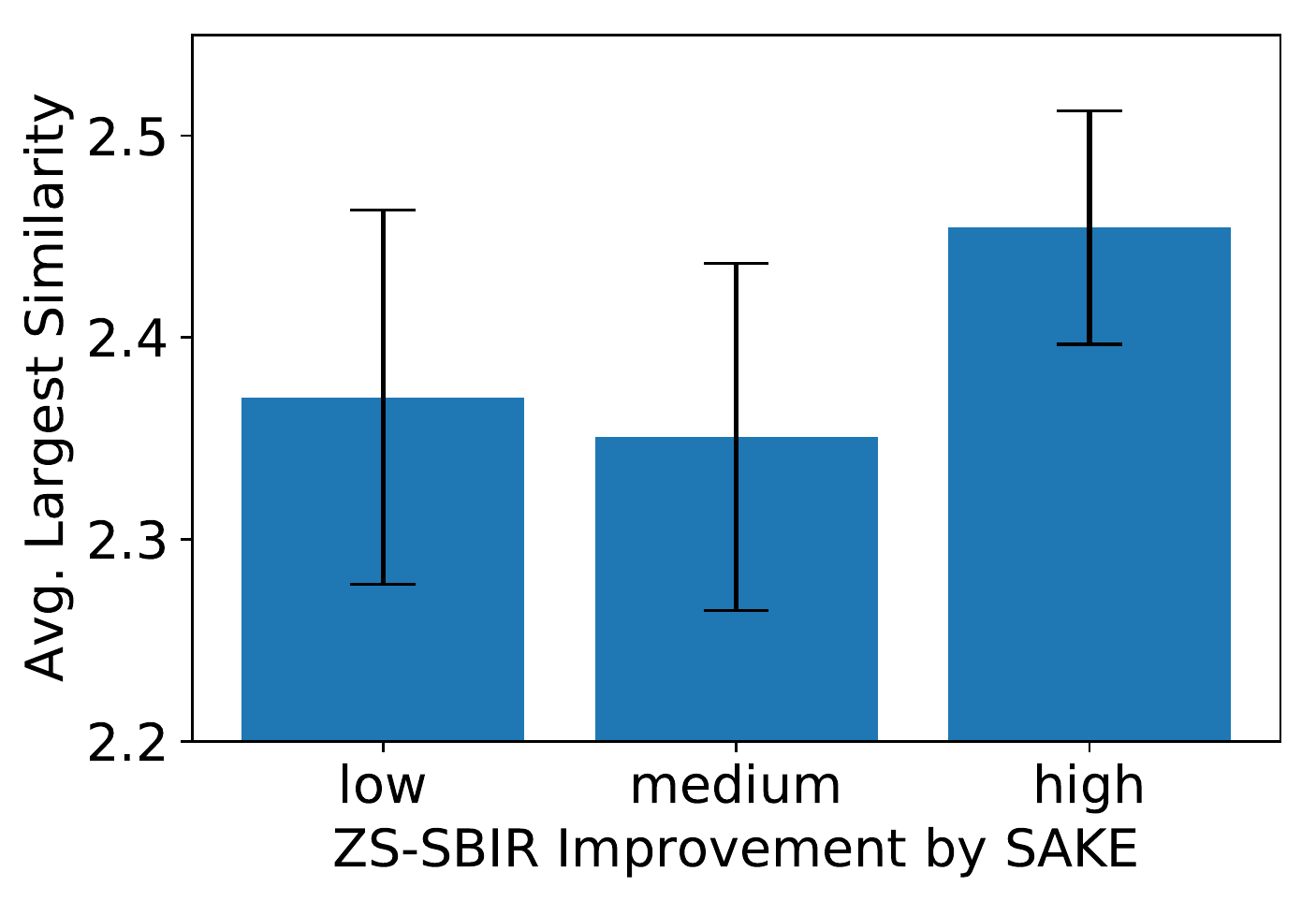}
%\vspace{-0.2cm}
\caption{ImageNet synset similarity with respect to the improvement brought by SAKE}
\label{fig:sm_synset}
\end{subfigure}
\vspace{-0.1cm}
\caption{How the ZS-SBIR mAP@all improvement brought by SAKE on different categories correlates to the category-level similarity to the original ImageNet categories. Error bars and mean values are summarized from three repeated experiments with different held-out category sets, all of which have no overlap with ImageNet.}
\label{Fig:sm}
\end{figure}

\subsection{Similarity by Classification Confidence}

In Figure~\ref{fig:sm_visual}, we can see that the improvement of SAKE becomes more significant when the category gets a higher classification confidence in an ImageNet-based classifier. This is to say, knowledge preservation, as expected, helps better to those categories that are closer to ImageNet -- in other words, these categories are often heavier impacted by catastrophic forgetting, and knowledge preservation brings more improvement.

We shall point out that this phenomenon does not mean that knowledge preservation is not useful to those categories which are not contained in the original domain. Indeed, in each of these 3 groups, we observe accuracy gain under the zero-shot setting -- this is also shown in our main experiments, in which both ImageNet and non-ImageNet categories largely benefit from knowledge preservation.

\subsection{Similarity by Semantic-space Distance}

To provide another perspective, we perform the same experiment using the similarity defined by the Leacock-Chodorow Similarity on the WordNet synset, {\em i.e.}, the semantic tree used to build up ImageNet. In Figure~\ref{fig:sm_synset}, we once again obtain the same conclusion, {\em i.e.}, SAKE is more effective on the categories that are better represented to the ImageNet semantic space.

\end{appendices}

\end{document}